# Lightweight Mobile Automated Assistant-to-physician for Global Lower-resource Areas

Chao Zhang[1,2], Hanxin Zhang[2], Atif Khan[2], Ted Kim[3], Olasubomi Omoleye[2], Oluwamayomikun Abiona[2], Amy Lehman[4], Christopher O. Olopade[3], Olufunmilayo I. Olopade[3], Pedro Lopes[3], Andrey Rzhetsky[2]

1. University of Chicago, Pritzker school of Molecular engineering
2. University of Chicago, Department of Medicine
3. University of Chicago, Department of Computer Science
4. Lake Tanganyika Floating Health Clinic

Emails: choozhang@uchicago.edu, hanxin@uchicago.edu, atifkhan@uchicago.edu, tedkim97@gmail.com, jimmyomoleye@gmail.com, oabiona@uchicago.edu, amy.ltfhc@gmail.com, solopade@bsd.uchicago.edu, folopade@medicine.bsd.uchicago.edu, pedrolopes@cs.uchicago.edu, arzhetsk@medicine.bsd.uchicago.edu.

## Key points

**Question**: Lower-resource areas like many countries in Africa are suffering from severe disease burdens, how can we utilize the modern health system in developed countries to help ease it?

**Findings**: Utilizing AI technology with large data sets from US, we developed a smart diagnosis assistance help primary healthcare providers in lower-resource areas document demographic and medical sign/symptom data and to record and share diagnostic data in real-time with a centralized database. The assistance system has been tested in Pakistan and proven to be effective.

**Meaning**: Our application would provide primary healthcare providers in lower-resource areas with a tool that enables faster and more accurate documentation of medical encounters.

## Abstract

**Importance**: Lower-resource areas in Africa and Asia face a unique set of healthcare challenges: the dual high burden of communicable and non-communicable diseases; a paucity of highly trained primary healthcare providers in both rural and densely populated urban areas; and a lack of reliable, inexpensive internet connections.

**Objective**: To address these challenges, we designed an artificial intelligence assistant to help primary healthcare providers in lower-resource areas document demographic and medical sign/symptom data and to record and share diagnostic data in real-time with a centralized database.

**Design**: We trained our system using multiple data sets, including US-based electronic medical records (EMRs) and open-source medical literature and developed an adaptive, general medical assistant system based on machine learning algorithms.

**Main outcomes and Measure**: The application collects basic information from patients and provides primary care providers with diagnoses and prescriptions suggestions. The application is unique from existing systems in that it covers a wide range of common diseases, signs, and medication typical in lower-resource countries; the application works with or without an active internet connection.

**Results**: We have built and implemented an adaptive learning system that assists trained primary care professionals by means of an Android smartphone application, which interacts with a central database and collects real-time data. The application has been tested by dozens of primary care providers.

**Conclusions and Relevance**: Our application would provide primary healthcare providers in lower-resource areas with a tool that enables faster and more accurate documentation of medical encounters. This application could be leveraged to automatically populate local or national EMR systems.

## Introduction

Diagnostic errors affect an estimated 12 million Americans each year [1]. There are multiple reasons for this large number; one reasons for this is physician fatigue and burnout. However, a deeper reason lies in the heavy-tailed pattern of disease frequency distribution [2]. Most diseases are in the very low prevalence area of overall disease distribution, thus, physicians who encounter too few instances of a complex pattern are less likely to recognize it. While this problem affects all physicians, it is exacerbated in resource-poor areas, where primary care providers function as generalists due to high demand and shortages of medical personnel.

Electronic Medical Records (EMRs) are rapidly growing in both volume and ubiquity, and this, along with the development of high-performance computing systems, is helping modern computers to outperform humans in many focused, diagnostic tasks. Attempts to develop efficient, computer-aided diagnosis support systems (DSSs) have been around since the 1960s [3-5]. While experimental DSSs that focus on single diseases have been successful in past decades, DSSs for general disease diagnoses have not flourished. There are a few general DSSs available commercially, like Isabel, DXplain, and GIDEON [6-8], but their clinical use remains limited due to their poor specificity and adaptiveness [9].

EMRs are considered essential to monitor and improve patient well-being [10]. Epic is the currently dominant clinical software company. Epic offers a clinical software suite that includes a communication portal and a set of specific diagnosis systems. Johnson et al. reviewed Epic's software comprehensively [11] and concluded that Epic provides a high-quality EMR system for collecting and managing accurate medical records. However, it comes with substantial costs: US physicians using Epic tend to complain about the work burden stemming from its module design, as well as the data entry involved to maximize billing, manual data coding, expensive, ongoing vendor support, and difficulty with introducing third-party extensions.

This inability of existing systems to meet the needs of lower-resource countries calls for a free, lightweight, network-free application that can perform documentation of patient encounters (creating EMRs) for general health problems and the tropical diseases that are prevalent in lower-resource settings – with both speed and accuracy. Our goal was to develop a system that would combine machine learning, computerized adaptive diagnosis [12,13], mobile application software, and EMRs for the accurate and easy capturing of patient-specific documentation across a broad spectrum of diseases, which would incorporate active-learning and functionality without internet connectivity. Paired with a centralized server which would receive messages from mobile terminals and dispense updated information to physicians, this system would provide a new Decision Support System (DSS) paradigm. If successful, this open-source software could be deployed globally.

To this end, we developed a computerized AI physician assistant system to aid primary care physicians in recording symptoms and diagnoses of a broad range of diseases, automatically recording patient encounters in a structured form. Learning from large US medical datasets, and using adaptive learning procedures and online learning algorithms, our system prompts primary healthcare providers with detailed, yet concise, suggestions of options to record symptoms, follow-up tests, and treatment options. Our light-weight mobile application uses negligible computational resources and is able to transfer data via text messages. Its intended users are primary healthcare providers in lower-resource areas with unstable internet connectivity.

The following sections discuss our system's design and current operation processes in greater detail.

## Materials and Methods

We designed the application with consideration to the disease burden in low-resource areas. Prior to the start of our system's development, we interviewed many primary care physicians in Nigeria, as summarized in Table 1. Most of the physicians still utilize hand-written patient records and think that a mobile application would ease their labor and improve effectiveness. Most of our interviewees preferred the Android platform. To increase user efficiency, our application works in an adaptive manner that requires only the minimum number of questions or tests to give suggestions, all based on natural language processing and decision tree algorithms.

This AI assistant is based on multiple data sets: Clinical notes from the University of Chicago Hospital, US MarketScan insurance claim data from IBM [14] (which includes electronic medical records for over 175 million unique patients and geographically from around the US), DeepDive medical literature open data sets from Stanford University [15] (which contains around 30 Gb of text data from open source medical literature and Wikipedia), the Unified Medical Language System (UMLS) at the NIH [16], and a knowledge database of disease-symptom associations generated with data from the New York Presbyterian Hospital, covering patients admitted during 2004[17]. We used the UMLS to define the medical terms, and the other four data sets to extract associations between those medical terms.

We used word2vec to process the medical literature and clinical notes [18]. word2vec is a natural language processing method that represents words as numerical vectors and learns the associations between words that appear closely in the corpus. As shown in Figure 1, we extracted the associations between medical terms and used them to construct a decision tree model [19,20].

By automatically learning association relationships from large corpora, word2vec is able to capture the association between words. We assume an association between words with a distance of less than 50 words, which is about the length of two sentences. The model remembers and learns the quantified co-existence relationship between medical terms. The advantage of word2vec is that it captures the contribution of synonyms. This is important because the manifestation of diseases and symptoms can vary greatly in the corpus.

After word2vec extracts the associations between symptoms and diseases, we constructed a decision tree to mimic the healthcare visit by generating questions iteratively based on previously collected information. From a symptom bank, it generates new questions by maximizing the information gain after the patient's initial information input or the last question. The information gain is maximized by finding the symptom item that separates the candidate diagnosis to two groups with similar size, which minimizes the statistical entropy.

In practice, patients are asked to provide their most concerning symptom as a trigger, which corresponds to the root of the decision tree. Next, by calculating the expected information gain for each symptom or lab test, the system adaptively decides what to ask in the next tree node. This process runs through several iterations and, when enough information is gathered, the system can then output diagnosis suggestions. This methodology minimizes the number of questions to ask or tests to perform and thus makes the diagnosis process much faster.

We developed a mobile application based on this model and further tested it in two hospitals in Pakistan. After collecting additional data, the application updates the parameters with a single-layer, artificial, neural network algorithm [21], which allows the assistant system to further improve as data accumulates. The entire code for the model and the Android application is open-source and can be found at https://github.com/imechaozhang/Mead-Android-App.

Figure 2 shows the mobile application's workflow. The application allows healthcare providers either to directly input both patient-specific symptoms and their diagnosis or ask for suggestions from the application. The application automatically saves everything locally in an SQLite database, which synchronizes with the remote server when internet is available. To optimize data storage scalability and accessibility, the remote server uses MongoDB, an open-source, document-based database. In order to function in lower-resource areas where internet connectivity may be unstable or non-existent, the application can communicate with a remote database via SMS messages. When healthcare providers choose to ask the application for diagnostic suggestions, the application will take the patient's basic information and their trigger symptoms and perform an adaptive diagnosis by asking about symptoms or tests with the largest information gain. Within a few steps, the application will provide some suggestions regarding diagnosis and prescriptions, as well as further, possibly necessary lab tests. Healthcare providers only need to click on a few buttons to complete the entire process.

While our application can make recommendations to healthcare providers from just a few clicks, that's all they are – suggestions. Healthcare providers are under no obligation to accept those suggestions, and indeed, a mechanism exists for them to reject the application's suggestions and type in their own diagnosis. We assume machine learning can never be 100 percent accurate and – importantly – our machine learning algorithm learns from mistakes.

## Results

We built, deployed, and tested an Android application intended to fulfill the need for decision-support systems to enhance physicians' diagnostic processes in lower-resource countries.

Figure 3 shows the application's key components accompanied by screenshots showing the application's workflow: basic information collection, symptom input, adaptive question generation, and diagnosis suggestion pages. There are also a few recording, confirmation, and orientation pages that guide the healthcare provider through the diagnosis process. The application has a few key functionalities: (1) Healthcare providers are able to visit and query the local database to pull up existing patient information, eliminating the need to input the basic information each time; (2) The application can work as an EMR system and physicians can simply input all the information gathered and save the information both on the Android device and in the Cloud (if internet is accessible); (3) The application's primary function is to achieve an adaptive diagnosis suggestion by generating questions regarding symptoms based on the information collected during the previous steps, and; 4) The application also considers lower-resource areas where internet connectivity is either unstable or unavailable by enabling synchronized data via encrypted SMS messages.

The current application's iteration relies on a "one-time pad" encryption scheme to protect information transferred over SMS. "One-time pad" is a symmetric encryption algorithm that uses a randomly generated key, known by only both the sender and receiver, to protect information security.

For the server, we used a Dell Optiplex 780 running Ubuntu 16.04 LTS. Our software is fairly hardware-agnostic, but we recommend that implementations based on ours use a 64-bit processor, with at least 250GB of storage memory, and at least 8GB of RAM. The device we used to test for SMS compatibility was a Huawei E8372 device, using an open-source python library to communicate with the device [https://github.com/pablo/huawei-modem-python-api-client]. The smartphones used for application development and testing were unlocked Samsung S5 models.

We beta-tested our initial Android application in Pakistan in early 2020 at Quaid-e-Azam International Hospital and Bashir Begum Memorial Welfare Hospital. We collected 113 diagnosis records for 99 individual patients, with 46 males and 53 females. Most patients were aged from 15 to 60, except two children and two seniors. During our testing, healthcare providers collected information from patients using the adaptive diagnosis process, in which the application sorted the disease list consisting of the 166 most common diseases and suggested five diagnosis results with the highest probability, given the information collected. Regardless of the suggestions from the application, the healthcare providers also made their own diagnoses. We considered the application's prediction to be accurate when the doctor's diagnosis lay within the application's top five suggestions. Figure 4 shows the application's diagnostic accuracy. The application suggested five diagnoses each time, and for 77 out of 113 cases, the doctors agreed with the suggestions from the application, and for 27 cases, they agreed with the first suggestion. The top-five accuracy comes to 68 percent and top-one accuracy is 24 percent. Note that this model is built out of medical literature without any real patient data, so there is much room for improvement.

While these application testing results are preliminary, they prove that our current prototype is functional and has the potential to serve a larger group of primary medical care providers. In our second development phase, we are making the application easier to use, more functional, and more accurate.

## Discussion and conclusions

Lower-resource countries are suffering from severe disease burdens and lack of well-trained healthcare providers. Current diagnosis systems are either not accurate enough or require too many resources, rendering them unusable in these markets. We are collaborating with the African Institute of

Mathematical Science and the University of Ibadan to design and deploy this electronic adaptive diagnosis system in order to help healthcare providers in Rwanda and Nigeria, as well as other parts of the world, such as Pakistan. This system is a lightweight, adaptive diagnosis assistant system designed to help healthcare providers arrive at diagnoses faster and with less errors and to assist with efficient patient record keeping. The adaptive diagnosis assistant system is based on online learning algorithms and will improve by itself as more patient data accumulates.

We developed this Android application prototype for use in clinics, and beta- tested it with five doctors in Pakistan. The performance and feedback were mostly positive, with an accuracy of 68 percent, and the doctors reported that the application was very user-friendly, and they liked the idea that it mimics a physicians' decision-making process. As to be expected, the doctor-to-patient ratio is extremely small in developing countries – doctors in rural or urban areas of Pakistan have huge patient loads with which they struggle. It is therefore important to provide a tool that can help improve their diagnostic efficiency and hasten their workflow, enabling them to potentially attend to more patients per unit time.

Our diagnostic application is specifically designed for lower-resource settings. While the application targets lower-resource settings, such as rural Nigeria, the US is not devoid of lower-resource areas. In the US, annually, over 12 million adults who seek outpatient medical care receive a misdiagnosis [22], and there nearly 30 million uninsured individuals (ten percent of the population) in the US. Our application may be able to help these people as well.

In lower-resource settings, particularly in countries ranking low on the Human Development Index, obtaining accurate and timely medical data is extraordinarily difficult. Without robust local datasets, the function and evaluation of our diagnosis system is limited in its present form. Although we have utilized large data sets from the US and experimented in Pakistan, it is unlikely that these data can properly represent rural African areas. We are in the process of deploying the application to countries like Nigeria to conduct a first trial experiment to collect more localized data, recognizing that there is vast heterogeneity across the African continent and also within countries -- especially between urban and rural contexts.

In the meantime, based on the feedback of doctors who tested the assistant system, it would be very useful if medical data could include more fields such as the duration and evolution of symptoms or patient medical histories. These are important factors that would help doctors arrive at better diagnoses and treatment plans. However, in developing this ideal system, we are restricted by both insufficient data sources and technology. The application requires a large amount of related data to add the requested features or medical fields to the system, and different algorithms are needed to handle the time-series information. We are planning to upgrade the system in this direction after the next trial in Nigeria.

Network accessibility may also be a problem when deploying the diagnosis system. To the best of our knowledge, cellular networks in many African rural areas are slow and brittle, and some rural areas have no cellular coverage at all. However, we did our best to build the lightest and most reliable diagnostic system possible. Consequently, the current version of the system only requires sending SMS messages of dozens of characters for each visit. Theoretically, this matches what we have seen so far in the field, but we have not tested the system while connected to an extremely unstable network; it is possible that data synchronization may fail. As we continue developing our AI assistant system, adding additional facets of patient health descriptors, the messages to be sent will grow larger, creating logistic difficulties in areas with low network capacity. We are reaching out to local universities and companies in Nigeria in search of testing support for these issues.

Here, we sincerely invite researchers working in clinical settings or data science to contribute to this open-source project in any format.

## Tables and Figures:

*Table 1: Survey results from physicians in Nigeria. Most of the physicians who took our surveys reported using hand-written recordings, which are not well-organized and sometimes even hard to read. They indicated they prefer to have an Android-based application, if provided. In the meantime, most of them think that a single mobile application would not be enough, reporting that the entire diagnosis and treatment lifecycle requires a full software suite.*

| Topic | Options | Responses | Comments |
|---|---|---|---|
| Preferred recording method | Hand-written | 43 | Many doctors (18) complained about hand-written recording |
|  | Digital | 6 |  |
|  | Total | 46 |  |
| Preferred electronic method, if provided | *Android* app | 39 | Some primary care providers already have access to both mobile and web apps |
|  | *iOS* app | 13 |  |
|  | Browser-based app | 22 |  |
|  | Total | 45 |  |
| Preference between single mobile | Single app | 6 | This is a follow-up survey and fewer doctors responded |
|  | Suite of apps | 10 |  |

| application and a suite of inter-connected apps | More than mobile apps | 6 | |
|---|---|---|---|
| | Total | 22 | |

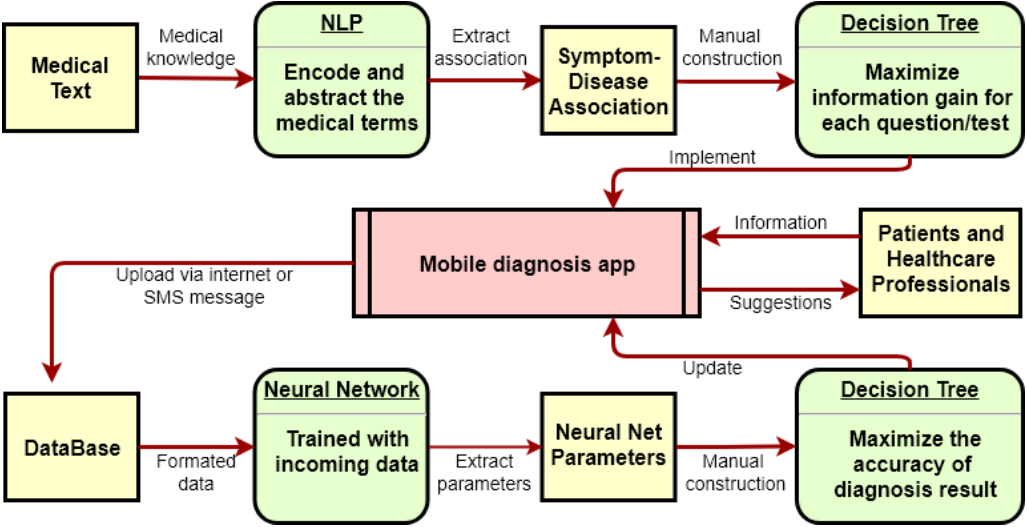

*Figure 1: DSS development workflow. Application `word2vec` extracts associations between medical terms (such as symptoms and diseases) from the medical text corpus. These associations are then used to construct a decision tree model for adaptive diagnosis, and the model is implemented into an Android application.*

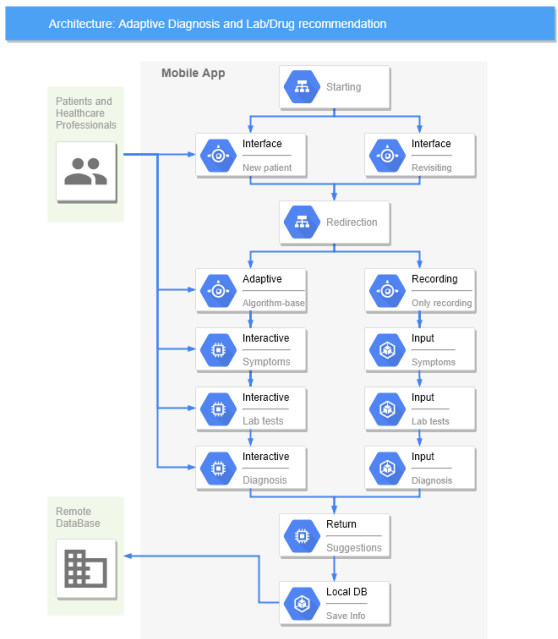

*Figure 2. The architecture of the mobile diagnosis application. It interacts with both patients and healthcare providers and optimizes the diagnostic process by an adaptive process based on a decision tree algorithm. All patient data are stored both locally and in the Cloud, so revisiting patients do not need to provide basic information. The data collected will be encrypted before sending to remote databases to ensure patient information security.*

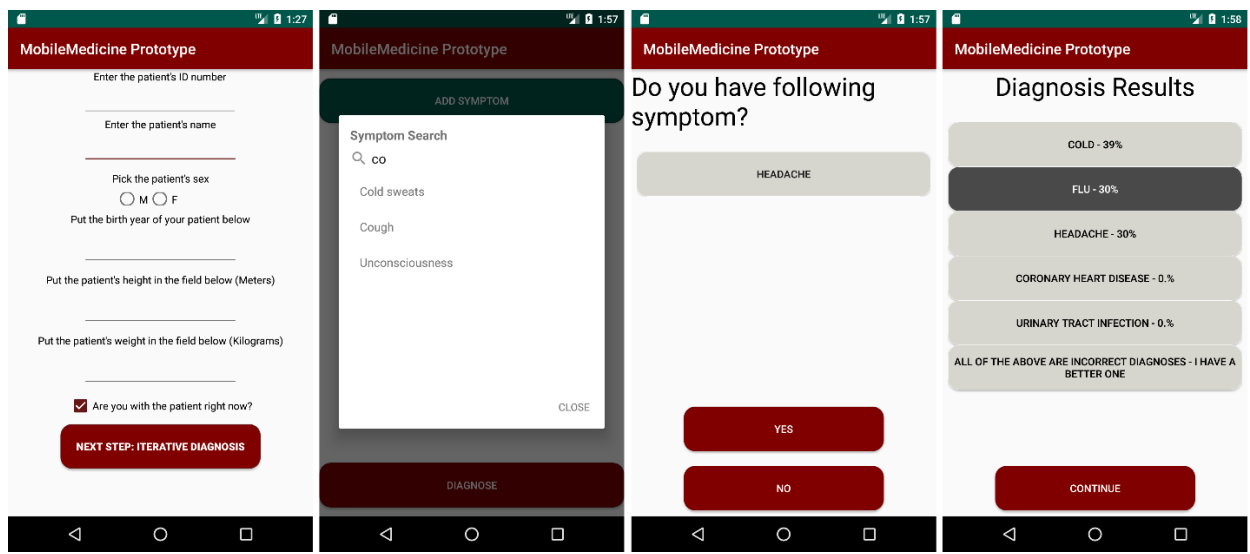

*Figure 3. Example screenshots from the Android application's prototype. 1) The basic information collection page, including the patient's age, gender, weight, and height. These are the most basic properties that are likely to be associated with disease.*

Other important information, such as family history or travel history, will be included in future versions. 2) The symptom input page, in which the primary care provider can select from the drop-down menu or type in patient symptoms. 3) The adaptive diagnosis page, where the application generates related questions about symptoms/tests and then generates new questions based on the answers. 4) The result page, which provides suggestions on diagnoses, tests, and prescriptions.

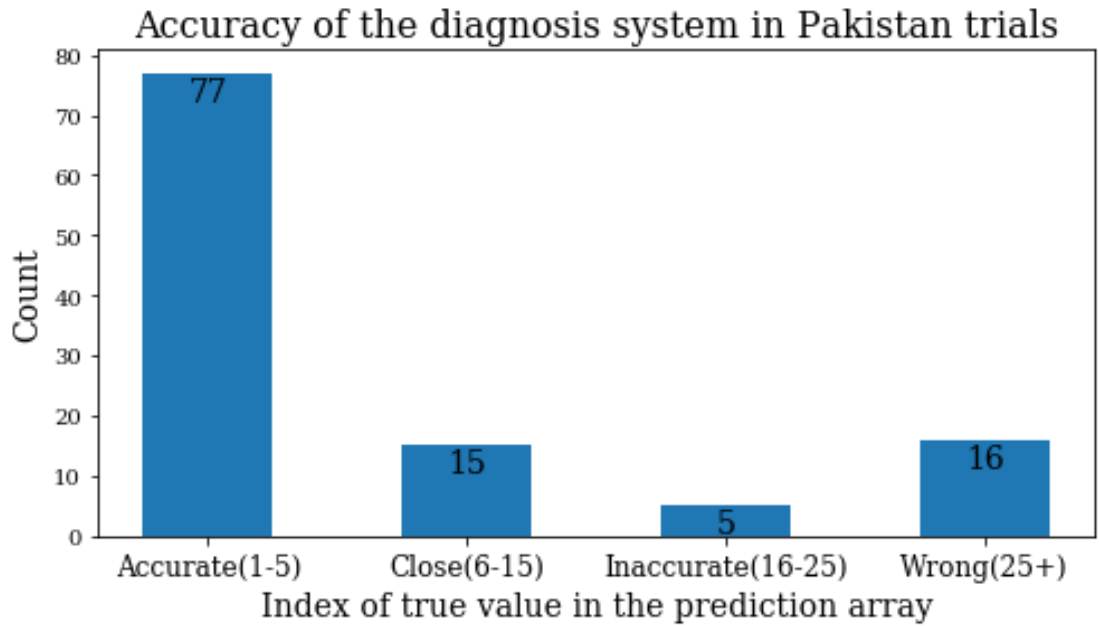

Figure 4. Results from the application's beta-testing in Pakistani hospitals. In this chart, the y axis is the total case count, while the x axis is the application's accuracy index based on healthcare provider diagnosis. The accuracy index is defined as following: The application covers the 166 most common diseases in the US and Africa. When used for diagnostics, we calculated the probability for all 166 diseases based on information collected from patients. The application sorted the disease lists according to the probabilities and suggested the top five diseases with the highest probability to healthcare providers, allowing them to choose from these five suggestions of diagnoses or reject them to make other diagnosis.